\documentclass[preprint,authoryear,12pt]{elsarticle}
\usepackage{amsmath,amsfonts,amssymb,epsfig}
\usepackage{subfigure}
\usepackage{graphicx}
\usepackage{fancybox}
\usepackage{color}
\usepackage{epstopdf} 

\def\etal{\textit{et al.\ }}

\newlength{\figureWidth}

\newcommand{\rev}[1]{\textcolor{black}{#1}} 
\newcommand{\revv}[1]{\textcolor{black}{#1}} 

\begin{document}

\title{Multi-Atlas Segmentation of Biomedical Images: \\ A Survey}
\author{Juan Eugenio Iglesias$^1$ and Mert R. Sabuncu$^2$}

\address{$^1$ Basque Center on Cognition, Brain and Language (BCBL), Spain. email: e.iglesias@bcbl.eu \\
$^2$ A.A. Martinos Center for Biomedical Imaging, Massachusetts General Hospital, Harvard Medical School, USA. email: msabuncu@nmr.mgh.harvard.edu}

\begin{abstract}
Multi-atlas segmentation (MAS), 
first introduced and popularized by the pioneering work of  \citet*{rohlfing2004evaluation},  \citet*{klein2005mindboggle}, and  \citet*{heckemann2006automatic}, is becoming one of the most widely-used and successful image segmentation techniques in biomedical applications.
By manipulating and utilizing the entire dataset of ``atlases'' (training images that have been previously labeled, e.g., manually by an expert), rather than some model-based average representation, MAS has the flexibility to better capture anatomical variation, thus offering superior segmentation accuracy. 
This benefit, however, typically comes at a high computational cost. 
Recent advancements in computer hardware and image processing software have been instrumental in addressing this challenge and facilitated the wide adoption of MAS. 
Today, MAS has come a long way 
and the approach includes a wide array of sophisticated algorithms that employ ideas from machine learning, probabilistic modeling, optimization, and computer vision, among other fields.
This paper presents a survey of published MAS algorithms and studies that have applied these methods to various biomedical problems. 
In writing this survey, we have three distinct aims.
Our primary goal is to document how MAS was originally conceived, later evolved, and now relates to alternative methods.
Second, this paper is intended to be a detailed reference of past research activity in MAS, which now spans over a decade (2003 - {\rev{2014}}) and entails novel methodological developments and application-specific solutions.
Finally, our goal is to also present a perspective on the future of MAS, which, we believe, will be one of the dominant approaches in biomedical image segmentation.
\end{abstract}


\maketitle

\section{Historical Introduction and Background}
Segmentation is one of the fundamental problems in biomedical image analysis and refers to the process of tagging image pixels or voxels with biologically meaningful labels, such as anatomical structures and tissue types. 
Depending on the application, these labels might constitute a handful of, possibly disjoint, regions of interest (ROIs) and a ``background'', which would refer to the parts of the image one might ignore in subsequent analysis. 
Alternatively, the labels might densely cover a substantial portion or all of the image, which is sometimes referred to as ``parcellation.'' 

The traditional approach to segment a given biomedical image involves the manual delineation (sometimes referred to as ``annotation'') of the ROIs by a trained expert.
This practice, however, can be painstakingly slow, prone to error, hard to reproduce, expensive, and unscalable.
Furthermore, the quality of the results will depend on the performance of the expert. 
Thus, manual delineation is typically not suitable for deploying on large-scale datasets or in 
applications where time is critical, such as treatment planning.
Automatic or semi-automatic segmentation algorithms can address these challenges, by speeding up the process, reducing the cost, and offering reliability, repeatability, and scalability.

Some segmentation algorithms, such as those that assign voxels to tissue types~\citep{kapur1996segmentation}, might not require the availability of training data in the form of manually delineated images (commonly called ``atlases'').
However, the class of methods we consider for this survey will depend on such training data and thus can be viewed as supervised learning algorithms. 
The goal of atlas-guided segmentation is to use/encode the relationship between the segmentation labels and image intensities observed in the atlases, in order to assign segmentation labels to the pixels or voxels of an unlabeled (i.e., novel) image.

In the early days of atlas-guided segmentation, atlases were rare commodities. 
In fact, in many applications, there was only a single atlas\footnote{The word ``atlas'' is a legacy of this era, where, for a given problem, one exploited a single map of labels denoting the biological meaning of the observed anatomy, for example, as captured by an image.}, i.e., a single image that was delineated by an expert.
In this context, the classical atlas-guided approach treats segmentation as an image registration problem~\citep{pham2000current}, where spatial correspondence is established between the atlas and novel image coordinates. 
Registration is typically a computationally expensive task that involves deforming (using some appropriate deformation model) one of the images until it is similar to the other one.
The resulting mapping between the two coordinate systems can then be employed to transfer (or ``propagate'') the segmentation labels from the atlas to the novel image voxels~\citep{christensen1997volumetric, collins1995automatic, davatzikos1996spatial, dawant1999automatic, lancaster1997automated, sandor1997surface}.
We refer to this technique as registration-based segmentation.

A single atlas coupled with a deformation model is usually insufficient to capture wide anatomical variation~\citep{doan2010effect}.
Therefore, the use of several atlases is expected to yield improved segmentation results.
Initial methods that utilized several atlases for segmentation took a two-step approach. 
In the first step, the most relevant atlas was identified, which was then used in a second registration-based segmentation step~\citep{rohlfing2003segmentation}. 
As we will see below, this can be viewed as a special case of multi-atlas segmentation, since all atlases are consulted for segmentation.
However, the approach that dominated early atlas-guided segmentation was probabilistic atlas-based segmentation~\citep{ashburner2005unified, fischl2002whole,park2003construction, pohl2006bayesian, thomas2008effects}, which had two distinctive properties. First, there was a single atlas coordinate frame, defined through the co-registration of the training images used to build the atlas. Second, statistics about the labels, such as the probability of observing a particular label at a given location, are precomputed in atlas space. The novel image was then segmented in the atlas coordinate frame with a probabilistic inference procedure that utilized \emph{parametric} statistical models. The spatial normalization  to the atlas could be computed via registration with a population template created at training, or estimated jointly with the segmentation within the probabilistic model; the latter alternative has the advantage that it is adaptive to variations in image intensity profiles, such as MRI contrast~\citep{ashburner2005unified}.

Probabilistic atlas-based segmentation offered two major advantages. 
First, by employing a single coordinate frame, to which all images were normalized, one automatically established spatial correspondence across all images. 
This facilitated the statistical analysis of biological variation across the population, as famously exemplified in voxel-based morphometry~\citep{ashburner2000voxel}.
The second advantage 
 was computational. 
One needed to run the computationally expensive image registration step (spatial normalization) only once per novel image. 

\begin{figure}[t!]
\setlength{\figureWidth}{0.9\linewidth}
\begin{minipage}{1.0\linewidth}
  \centering
  \centerline{\epsfig{figure=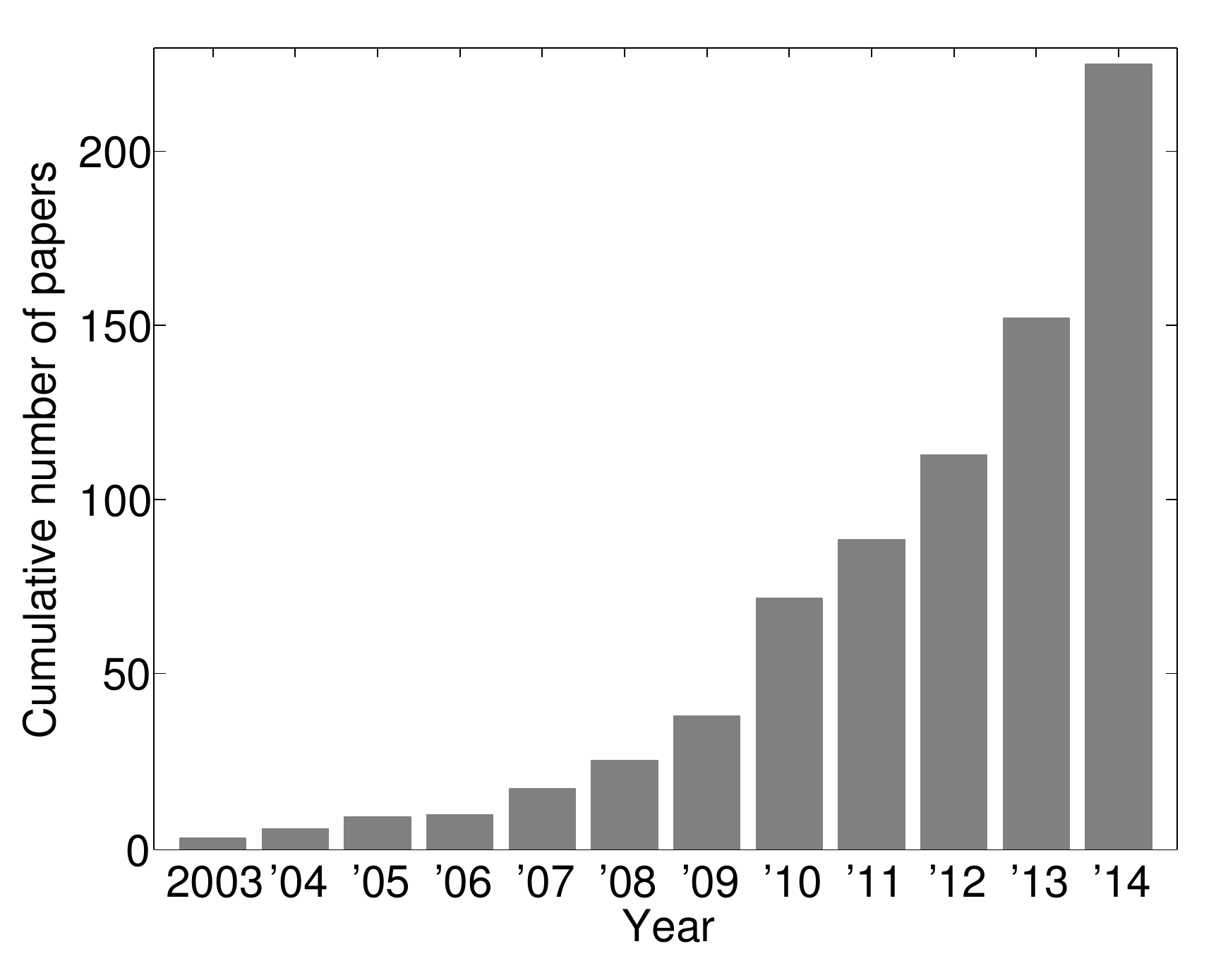,width = \figureWidth}}
\end{minipage}
\caption{Cumulative number of papers, cited in this survey, that introduce a novel MAS method or present a novel application of MAS.}
\label{fig:cnt}
\end{figure}

In 2003-2004, in a series of papers~\citep{rohlfing2003expectation,rohlfing2003expectation2,rohlfing2003extraction, rohlfing2004evaluation}, Rohlfing and colleagues proposed an alternative segmentation strategy, which at the time might have not seemed radically different.
Yet, as we elaborate below, this work inspired a rapidly growing class of methods (see Figure~\ref{fig:cnt}), including the pioneering work of \citet{klein2005mindboggle}, \citet{heckemann2006automatic} and others.
We collectively refer to these methods as \emph{multi-atlas segmentation} (MAS).
In this approach, the atlases are \textit{not} summarized in a (probabilistic) model.
Instead, each atlas is available and potentially used for segmenting the novel image.
A classical example involves applying a pairwise registration between the novel image and each atlas image.
These registration results are then used to propagate the atlas labels to the novel image coordinates, where at each voxel, the most frequent label is selected.
This is commonly referred to as ``majority voting.''

\begin{figure}[t!]
\setlength{\figureWidth}{0.9\linewidth}
\begin{minipage}{1.0\linewidth}
  \centering
  \centerline{\epsfig{figure=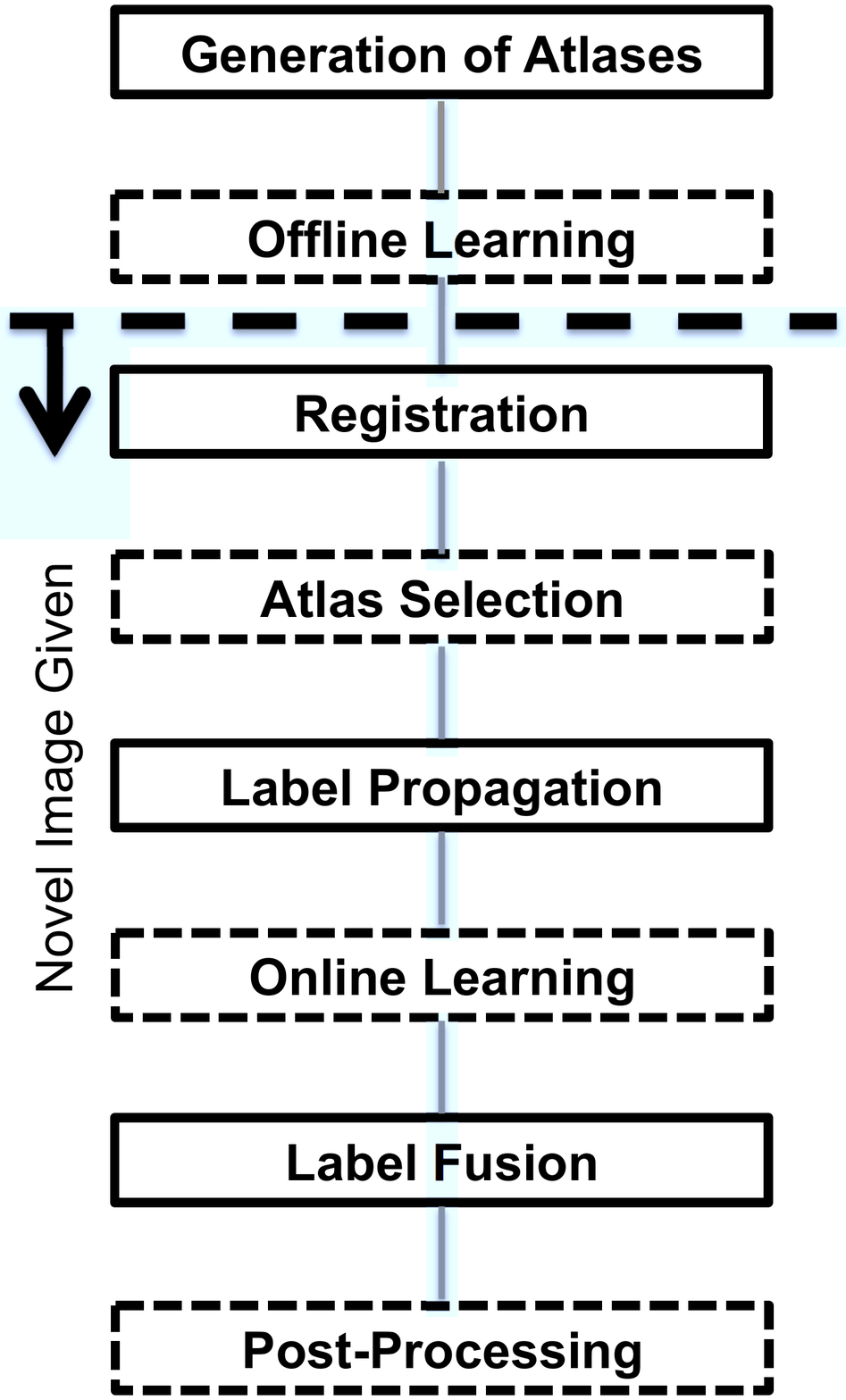,width = \figureWidth}}
\end{minipage}
\caption{Building blocks of MAS. Dashed blocks can be considered optional.}
\label{fig:flow}
\end{figure}

We can subdivide a MAS algorithm into several components that we depict in Figure~\ref{fig:flow}.
These components might be implemented as independent, sequential steps, where earlier steps are placed above in the illustration.
However, there are many exceptions to this structural organization.
For example, in some algorithms, some blocks might be unified, form feedback loops, swap places, or even be omitted altogether.  
That said, we find this diagram useful for organizing methodological developments in MAS.
Therefore, the part of our survey covering methods will adhere to this organization, with subsections corresponding to each one of these components.
 
 The remainder of this survey is structured as follows.
 Section~\ref{sec:methods} presents an account of published MAS methods, organized into the aforementioned building blocks.
 Section~\ref{sec:app} surveys published studies that apply a MAS algorithm to a novel biomedical problem.
We conclude with a discussion and pointers to promising future directions of research in Section~\ref{sec:disc}.
Finally, we would like to note that we have made all effort to cover the relevant literature as comprehensively as possible (\rev{as of the end of 2014}).
Yet, we are bound to have missed some pertinent publications.
Furthermore, we made the conscious choice to leave out some redundant papers. 
For example, earlier conference versions of more detailed journal publications were typically omitted.
\section{Survey of Methodological Developments}
\label{sec:methods}

\subsection{Generation of Atlases}
\label{sec:genAtl}
Atlases, i.e., labeled training images, form the core foundation of MAS algorithms.
They are typically obtained by the meticulous and costly effort of a domain-specific expert who relies on an interactive visualization software, such as~\citep{criminisi2008geos,heiberg2010design,pieper20043d, yushkevich2006user}, and might exploit multiple imaging modalities, while recruiting textbook anatomical knowledge. 
However, as we discuss below, there are exceptions to this rule.

Before seeing the to-be-segmented novel image, most methods treat each manually segmented image equally. 
Yet, to improve performance, one might identify high quality training cases, \rev{for example} via visual inspection~\citep{yang2010automatic}. \rev{Another option is to borrow ideas from the feature selection literature~\citep{pudil1994floating} to automatically preselect the subset of atlases that is expected to yield the maximal performance when new data are segmented~\citep{iglesias2009robust, isgum2009multi}. Prior knowledge on the problem at hand can also be used to preselect the atlases. For example, \citet{tung2013multi} selected the atlases with narrow lumen in a neointima segmentation method in coronary optical coherence tomography (OCT), since neointima only exists in coronary arteries with narrow lumen.  Such approaches can increase the accuracy of segmentation by discarding low quality training data, but a decrease in segmentation quality is also possible due to the reduction of the atlas pool size, which also represents a waste of labeling effort. Another way of improving the performance of the segmentation is} to apply population-level preprocessing \rev{(e.g., by co-registering the atlases)} to increase their signal-to-noise (SNR) ratio~\citep{zhuang2010whole}. \rev{While higher SNR can yield a more accurate registration and segmentation of the novel image, this improvement will depend on the quality of the co-registration and noise properties of the data. A similar approach was recently used to compute multiple population templates via a clustering strategy, which were manually labeled~\citep{gao2014multi}.}

In applications where the atlases might not be a representative sample of the population, one can synthesize atlases that offer a better representation of anatomical variability. \rev{Despite the higher computational cost in the subsequent analysis, such an approach can increase the accuracy of the segmentation by enriching the atlas pool. For example, ~\citet{jia2012iterative} used a statistical model based on principal component analysis (PCA) to synthesize deformations. One disadvantage of this technique is that the synthetic deformations might not always be anatomically plausible.}
In a related effort, ~\citet{doshi2013multi} proposed to cluster all available training images \rev{using k-means on the L2 norm of intensity differences in order} to identify a representative subset of cases that can be then manually labeled. \rev{Their algorithm, which is closely related to active learning, can greatly reduce the manual labeling effort, but also requires a large pool of unlabeled data.} 
Recently, \citet{awate2014multiatlas} presented a strategy that used a small number of labeled cases \rev{and a model of MAS based on non-parametric regression in the space of images, in order} to predict the total number of atlases that need to be manually segmented to obtain a desired level of segmentation accuracy within a MAS framework. \rev{This technique can be useful for planning the manual segmentation phase.}

In an alternative approach, one might exploit the wide availability of non-expert segmenters, instead of trying to obtain high quality expert manual segmentations. \rev{Trading off quality of annotations for number of atlases can be beneficial in some applications.}
For example, \citet{bogovic2013approaching} consider this scenario and propose to directly model the unknown ``expertise'' of each atlas, \rev{encoded in a confusion matrix between true and estimated labels.} \Citet{bryan2014self}, on the other hand, consider relying on the self-declared ``confidences'' of the manual segmenters, \rev{which are used to weigh their contributions when merging their ``opinions'' in the segmentation of novel images.} 
In other scenarios, atlases might have been segmented multiple times, as in~\citep{weisenfeld2011softstaple,wang2012guiding}, or only portions (e.g. certain slices) of the training data might have been manually traced, as in~\citep{landman2012robust}. \rev{This information needs to be considered in the subsequent steps of the MAS pipeline.} 
Finally, there are several proposed approaches, e.g.~\citep{chakravarty2013performing, gass2013semi, heckemann2006automatic,jia2012iterative, liao2013sparse, shen2010supervised, wang2013groupwise, wolz2010leap,kotrotsou2014ex,janes2014striatal}, that exploit the novel, unlabeled images to enrich the training data, for instance, by employing automatic segmentations as atlases, \rev{or by using them to generate different registration paths between the atlases and the target scan to segment. The former can be seen as a form of self-training (a semi-supervised learning technique in which unlabeled data are automatically classified and added to the pool of training samples). While it can take advantage of unlabeled data, it also inherits self-training's shortcoming that segmentation mistakes reinforce themselves. The latter (using multiple registration paths) can increase the performance by generating more training segmentations for label fusion (i.e., the combination of propagated labels into a segmentation estimate), but at the same time, these might be poor candidates due to suboptimal registration and eventually worsen the final segmentation.}

\subsection{Offline Learning}

Classical MAS algorithms applied no or very little processing to the atlas data \emph{offline}, i.e., prior to observing the novel image.
Atlases were manipulated and analyzed solely based on information from the image to be segmented.
However, some of the more recent methods we review here perform what we call ``offline learning,'' where the atlases are analyzed offline and some sort of information is garnered to be used during the segmentation of the novel image.
For example, one can learn a strategy to compute rough regions of interest in the novel image, in order to constrain or guide subsequent processing steps \rev{and/or reduce computational cost}~\citep{li2013regression,ramus2010construction,van2010adaptive, shi2010construction}.
In a very different approach, \citet{van2008hippocampus} proposed to construct a likelihood model on the training data, which quantifies the probability of observed image intensities conditioned on the underlying labels. \rev{While such a model can improve the segmentation by linking labels and image intensities, it can also degrade it if the intensities of the atlases and novel image are not well matched.} Similarly, \citet{zikic2014encoding} suggest to train a \rev{random forest} classifier corresponding to each atlas, which learns to predict labels based on the image appearance. Instead of labels propagated via a registration step, atlas predictions computed using these classifiers are combined into a segmentation. \rev{Since classification with random forests is fast, this method can be computationally more efficient than conventional registration-based multi-atlas segmentation, but also less accurate, as it will fail to capitalize on the high accuracy of modern registration methods.}
In a related effort, \citet{wang2013multi2} considered a tumor segmentation application, where the algorithm cannot rely on spatial correspondences between the images. Instead, they employed a data-driven clustering strategy on atlas voxels to identify super-voxels (i.e., patches of irregular size), which were then used by a k-nearest neighbor classifier to segment the novel image.
\rev{In a parallel effort, \citet{wang2014multiLBLF} proposed to use a local random forest classifier trained on the atlases to predict the segmentation label in the novel image.} 

Another
 direction involves analyzing the training data in order to learn how to assign weights to each atlas when conducting label fusion. 
One such strategy estimates measures of reliability associated with \rev{the atlases by co-registering them and computing the agreement between the propagated labels; atlases than can better predict the labels of others' receive higher weights}~\citep{sdika2010combining,wan2008automated}.
Alternatively, supervised learning approaches have been proposed to predict  the weights from the novel image. \rev{For example, \citet{sjoberg2013multi} pre-registered the atlases and learned the distribution of Dice scores given the registration similarity measure; label fusion weights were derived from this distribution. In a similar fashion, \citet{sanroma2014learning} trained a support vector machine to predict a ranking of the atlases based on image features.} \rev{The main disadvantage of these approaches is that they do not always generalize well beyond the training data.}
A related, yet different technique involves applying clustering~\citep{langerak2013multiatlas,shi2010construction}, manifold learning~\citep{cao2011putting, cao2011segmenting,cao2012multi, duc2013using, wolz2010leap, gao2014multi}, or computing a minimum spanning tree on the atlases~\citep{jia2012iterative}. 
These learning algorithms are employed to construct a structure on the space of training images, which yields the means to efficiently
compute distances between the atlases and novel image(s), run registrations, and propagate manual labels.

\subsection{Registration}
\label{sec:registration}
Registration is the task of establishing spatial correspondence between images and is considered one of the fundamental problems in biomedical image processing.  
Image registration involves deforming (or warping) one or more images to \rev{maximize an objective function that combines a metric of spatial alignment with a regularizer that quantifies the plausibility of the deformation.}
\rev{The three components of an image registration algorithm are thus the deformation model, the objective function, and the optimizer. The deformation model represents the class of spatial transformations that are plausible in the application at hand. This can be as simple as a rigid transform, or as complex as a non-parametric model in which each location is assigned a spatial transformation vector. Some deformation models incorporate constraints that exploit prior knowledge to make the spatial transforms more realistic. These constraints can be integrated in the deformation model (e.g., inverse consistency, diffeomorphism), or explicitly specified in the objective function through regularizers. The objective function is typically based on either the spatial distance between corresponding landmarks (manually placed or automatically detected) or on image intensities. In the latter case, metrics such as sum of squares or cross-correlation have been widely used in intramodality scenarios, whereas statistical metrics such as mutual information have been popular when registering across modalities. Finally, the optimization method is often an iterative algorithm, e.g., gradient descent, conjugate gradient, Levenberg-Marquardt, of the BFGS algorithm. However, discrete, graph-based methods are also becoming popular, e.g.,~\cite{glocker2008dense}. An extensive review on deformation models, objective functions and optimizers can be found in \citet{sotiras2013deformable}.} 

The optimal choice of algorithm specifics largely depends on the biomedical application, its goal~\citep{yeo2010learning}, and operational constraints, such as available computational resources, desired accuracy, and restrictions on time. 
\rev{Once registration is complete,} the resulting spatial transform can then be used to map from the frame of one image to the coordinates of another.

In MAS, registration is the step that determines the spatial correspondence between each atlas and the novel image.
Early MAS methods, such as~\citep{heckemann2006automatic, rohlfing2004evaluation,wan2008automated}, relied on nonlinear deformation models, such as spline-based parameterized transformations~\citep{rohde2003adaptive, rueckert1999nonrigid} or non-parametric diffeomorphisms~\citep{beg2005computing,vercauteren2009diffeomorphic}, which seek voxel-level alignment accuracy. 
Several studies~\citep{bai2012atlas,lotjonen2009atlas,lotjonen2010fast,sjoberg2013multi} have conducted empirical comparisons of the impact of different registration algorithms on MAS performance in different applications.

Typically, \rev{one independent} registration is computed between each atlas and the novel image and generic intensity-based registration tools, such as~\citep{avants2009advanced,klein2010elastix,rueckert1999nonrigid,ou2011dramms}, are used.
Yet, \citet{rohlfing2005multi} experimented with running the registration step several times with different parameter settings and combining all resulting propagated labels. \rev{As explained in Section~\ref{sec:genAtl} above, generating more candidate segmentations can improve the subsequent fusion, but also might worsen it by introducing poor candidates generated by suboptimal registrations}.
A similar strategy, proposed by \citet{wang2013multi3}, employs precomputed registrations between pairs of atlases to generate a multitude of propagated labels by concatenating the pairwise registration results. 
 \rev{In a parallel effort, \citet{datteri2014applying} relied on pre-registered atlases to estimate registration accuracy for the novel image based on registration circuits.}
Also, several authors proposed to employ the manual segmentations~\citep{han2008atlas,nie2013automated,tamez2012unsupervised,lee2014multi}, multiple imaging channels~\citep{yushkevich2010nearly}, or automatically computed tissue maps~\citep{heckemann2010improving,ledig2014robust} to establish more accurate alignment, \rev{ often at a higher computational cost}.
\rev{Instead of computing the pairwise atlas-to-novel image registrations independently, \citet{lee2014multi} recently proposed to solve them simultaneously in a group-wise registration framework}.
In another parallel effort, motivated by the observation that the registration step would benefit from the knowledge of the underlying segmentation labels, \citet{hao2012iterative}, \citet{iglesias2013unified}, \citet{tang2013bayesian}, \rev {and \citet{stavros2014discrete}} developed MAS algorithms that integrated the registration and label fusion steps. Thus, instead of treating registration as an independent preprocessing step, these algorithms iterate between registration and segmentation, \rev{which yields a small increase in segmentation accuracy at the expense of reduced computational efficiency}.

Typically, MAS treats the unknown deformation between the atlas and novel image as a nuisance, which once computed is only used to deform the atlas image intensities and/or propagate the labels.
Yet, a growing number of methods recognize the value in the deformation fields themselves and propose to use information about the amount of deformation in \rev{the computations of the fusion weights. For example,  \citet{commowick2007efficient} used the Euclidean norm of the deformation, \citet{ramus2010construction} used its Jacobian determinant, and \citet{wang2014multiAutoSeg} used its harmonic energy}.

The registration step is the computational bottleneck of the MAS algorithm and largely determines run time.
One strategy to reduce the computational burden introduced by registration is via atlas selection (see next section), which can obviate expensive registrations with unselected atlases.  
An alternative, popular approach employs a common coordinate system, similar to conventional probabilistic atlas segmentation methods, either via a standard template~\citep{aljabar2007classifier,aljabar2009multi}, a population average~\citep{artaechevarria2008efficient, commowick2007efficient, commowick2009using, depa2011towards, fonov2012multi, ramus2010construction, shi2010construction, shi2013conformal, zhuang2010whole}, or one of the atlases~\citep{van2010adaptive,sjoberg2014much}.
Here, all atlases are co-registered offline, and the novel image is registered with the template image. 
The template-to-novel image transformation can then be concatenated with the atlas-to-template transformations in order to propagate labels from the atlases to the novel image~\citep{artaechevarria2008efficient,depa2011towards, ramus2010construction,rivest2014fast}. \rev{Such an approach can reduce the computational cost of registration, but also might negatively impact the performance due to the suboptimality of the registrations.}
The use of a common coordinate frame further enables the definition of regions of interests, which the segmentation algorithm can employ in subsequent steps, \rev{e.g., atlas selection~\citep{shi2010construction} or label fusion~\citep{commowick2009using,ramus2010construction}.}
Yet another strategy to accelerate the registration step is to exploit the rapidly growing availability and capability of GPU processors, as proposed in~\citep{duc2013using,han2009gpu,modat2010fast,cardoso2013steps}.

A recent technique is inspired by the non-local means method~\citep{buades2005non} and utilizes a patch-based search strategy to identify correspondences with the atlases.
This technique was introduced to biomedical MAS by \citet{coupe2011patch} and recently has been gaining popularity~\citep{asman2013non, bai2013probabilistic,fonov2012multi, konukoglu2013neighbourhood, rousseau2011supervised, wang2011regression, wang2013multi, wolz2013automated, wang2014segmentation, zhang2011confidence, zhang2012sparse,wang2014automated,wang2014multiLBLF,ta2014optimized, wang2014geodesic,wang2014integration,sanroma2014novel}.
These papers have demonstrated that a patch-based search strategy can be used in a wide range of MAS methods to improve performance, for example, by relaxing the one-to-one correspondence assumption or eliminating the need for highly accurate registration results. 
\rev{In contrast with common implementations of non-local means in computer vision, its biomedical applications can be computationally efficient, cf.~\citep{ta2014optimized}, for instance, by assuming a rough alignment (e.g., achieved via a linear transformation model), which allows one to restrict the patch search to a local neighborhood of each voxel. Furthermore, the anatomical context can be used to improve the quality of the patch matches, as demonstrated in~\citep{wang2014geodesic,wang2014integration}.}

\subsection{Atlas Selection}

There are two main motivations not to use all available atlases in MAS.
First, by reducing the number of atlases, one can improve computational efficiency. 
This might be particularly important for applications where time is a significant constraint.
A typical MAS algorithm's computational demand is at least linear with respect to the number of utilized atlases. 
So, selecting only half of all available atlases would be expected to about double the speed of the algorithm and reduce the memory requirements by up to a half.  
Second, by excluding irrelevant atlases that might misguide the segmentation procedure, one might expect to improve final segmentation accuracy.
The specifics of the problem and utilized algorithm determine how applicable and significant these two points are.
For example, it has been observed that atlas selection can improve the accuracy of majority voting~\citep{aljabar2007classifier}, but is less critical for weighted fusion~\citep{sabuncu2010generative}.

\rev{The effectiveness and efficiency of atlas selection are closely related to the registration step. If all atlases are nonlinearly registered to the novel images, the atlas selection step will be well informed by the similarity of the images, but the computational savings will be nonexistent or minimal. If no registration is performed, the computational efficiency of the algorithm is much higher, but it is more difficult to select the atlases that are most relevant to the novel images to segment. A compromise can be achieved by linearly registering all the atlases to the novel scan, performing the selection, and continuing with the nonlinear component of the registration only for the chosen atlases.}
\rev{While the selection is typically conducted prior to the segmentation of the novel image, several authors~\citep{langerak2013multiatlas, langerak2010label, van2010adaptive, weisenfeld2011learning} have proposed methods that iterate between segmentation and atlas selection, pruning or adding to the selected atlas set based on the current estimate of the segmentation. }

Early atlas selection methods employed a metric to rank the relevance of the atlases.
These metrics included similarity measures based on image intensities, e.g., sum of squared differences, correlation or mutual information~\citep{aljabar2007classifier, aljabar2009multi, aribisala2013assessing, tung2013multi, xie2014low, wu2007optimum}; non-image meta-data such as age~\citep{aljabar2007classifier, aribisala2013assessing}; registration consistency~\citep{heckemann2009mirror}; amount of deformation~\citep{commowick2007efficient, commowick2009using}; and anatomical geometry~\citep{teng2010head}. 
Several studies have conducted empirical comparisons of these different atlas selection strategies in various MAS applications~\citep{aljabar2007classifier,acosta2011evaluation, avants2010optimal, lotjonen2010fast, ramus2010assessing}\rev{, concluding that sum of squared differences and cross-correlation (after histogram matching) of intensity values, along with age difference, are reliable metrics to rank the relevance of atlases}.

\rev{More recently,} other works have proposed to define similarity measures based on an image manifold structure.
\rev{For example, \citet{duc2013using} used Isomap, locally linear embedding and Laplacian eigenmaps to learn the manifold. \citet{cao2011segmenting} used locality preserving projections, and \citet{asman2014groupwise} used principal components analysis. These approaches introduce additional complexity to the system, but can outperform standard similarity measures in the atlas selection task.} 
Another approach to increase the efficiency and accuracy of atlas selection utilizes clustering, where the atlases, possibly together with the novel image(s), are analyzed to identify clusters of similar cases \rev{using methods such as k-means~\citep{nouranian2014multi}, affinity propagation~\citep{langerak2013multiatlas} and Floyd's algorithm~\citep{wang2014multiAutoSeg}.} 
Then, cluster representatives (or exemplars) are used for the initial search of the most relevant atlases. 
\rev{These methods can yield a performance similar to approaches that do not preselect atlases, but at a much lower computational cost.}
Alternatively, atlas selection can be treated as a learning problem, where the optimal strategy to choose the relevant atlases can be learned on the atlases themselves, utilizing the manual segmentations, \rev{as demonstrated by \citet{cao2012multi} (manifold learning), \citet{konukoglu2013neighbourhood} (random forests) and \citet{sanroma2014learning} (support vector machines). These strategies are not straightforward to implement but have been shown to improve segmentation performance.} 

While the atlas selection method 
has a significant impact on segmentation performance, with notable exceptions~\citep{awate2014multiatlas, heckemann2006automatic}, the optimal number of atlases to be selected seems to be an overlooked topic of research.
Some algorithms simply choose the most suitable single atlas, and apply registration-based segmentation~\citep{commowick2007efficient, teng2010head, wu2007optimum}. 
Yet most MAS methods end up using more than one atlas~\citep{rohlfing2004evaluation,klein2005mindboggle,heckemann2006automatic}. 
Typically, the number of atlases to be selected is either estimated, e.g., based on heuristics such as computational expectations, or determined empirically via cross-validation, bootstrapping or a similar sampling strategy.

\subsection{Label Propagation}

Once the relevant atlases are selected, and spatial correspondence is established with the novel image, the classical multi-atlas segmentation strategy proceeds by propagating the atlas labels to the novel image coordinates. 
Since early MAS methods~\citep{heckemann2006automatic}, one of the most popular strategies has been to utilize ``nearest neighbor interpolation,'' where each atlas transfers a single label to each novel image voxel, 
e.g.,~\citep{artaechevarria2009combination, langerak2010label, langerak2013multiatlas,sabuncu2010generative}. 
Although this label
corresponds to that of the closest voxel in atlas space, 
\rev{\citet{sdika2010combining} showed that higher performance can be achieved by augmenting the information with a tissue consistency step.} 
That is, the nearest neighbor search is conducted among those atlas voxels with a tissue segmentation (obtained automatically, from a separate step) consistent with the target voxel.
\rev{However, this approach depends on the performance of the separate tissue segmentation step, which can be sensitive to outliers, as in the case of very old or diseased subjects not represented in the atlas pool.}
The nearest neighbor strategy can further be 
 refined using, for example, linear interpolation~\citep{rohlfing2004evaluation,sabuncu2010generative}, where each atlas's vote is spread over multiple labels, with associated weights that reflect the ratio of partial volumes.

An alternative approach involves using the signed distance maps of the original atlas label images~\citep{gholipour2012multi,gorthi2013weighted,sabuncu2010generative, sjoberg2013multi,weisenfeld2011learning,xu2014shape}. 
Each label has an associated signed distance map, which takes positive values within the corresponding structure, negative values outside, and the magnitude is proportional to the closest distance to the label boundary.
The signed distance map
encodes the uncertainty close to label boundaries and the relative confidence deep within a region. 
While signed distance maps are not naturally normalized (i.e., the scale depends on the size and shape of the anatomical structure), one strategy is to use them to compute label probabilities, e.g., via the logistic mapping~\citep{sabuncu2010generative}.
A complementary technique transforms the atlas label boundaries directly, rather than applying a volumetric warp to the images~\citep{chou2008automated, klein2005mindboggle, nie2013automated, tamez2012unsupervised}.
Finally, rather than transferring over atlas labels via a geometric deformation model, one can employ learning algorithms trained on each atlas to generate voxel-level candidate label estimates for each atlas, as recently proposed in~\citep{zikic2014encoding}. \rev{This technique does not seem to increase segmentation accuracy, but can considerably reduce the run time of the algorithm by requiring only one nonlinear registration (to align a probabilistic atlas to the novel scan).}

\subsection{Online Learning}

The labels of the atlases that have been propagated to novel image coordinates are often merged directly into a single estimate of the segmentation with a label fusion algorithm. 
However, several MAS methods perform an ``online learning'' step, which aims to boost the performance of the algorithm by exploiting the relationships between the registered atlases and the novel images. 

Some methods use the estimated segmentation of the novel image to iteratively perform atlas selection and/or registration. 
For example, the selected atlas set can be determined based on the similarity between the deformed atlas labels and the current estimate of the segmentation, \rev{which can increase segmentation accuracy by excluding outlier atlases from fusion}~\citep{langerak2010label}.
Alternatively, \citet{van2010adaptive} divide the novel image into blocks, which are used to update the local registrations and selection of atlases. \rev{This way, they are able to automatically stop the local registration of atlases when no improvement is expected, reducing the computational cost without a negative impact on performance.}

Other approaches \rev{exploit} the relationship between the labels and intensities of the novel image in order to assist the fusion step.
This can be achieved via conditional Gaussian models~\citep{lotjonen2009atlas}, or non-parametric density estimators~\citep{weisenfeld2011learning}, which can be employed to refine the propagated labels. \rev{However, this strategy can be counterproductive when the intensities of the atlases and the novel scan are not well matched.}
In a related effort, \citet{hao2014local} use a discriminative technique to model the posterior label probability directly. 
\rev{More specifically, they use the intensities and labels of the registered atlases in a window around each spatial location to build a set of local classifiers -- one per voxel of the novel image. Each classifier is an L1-regularized support vector machine that predicts the label of the voxel at hand from hundreds of image features computed with different filters. In this case, label fusion is implicitly carried out in the classification.}

Finally, we have semi-supervised approaches that utilize the collection of novel images along with the atlases.
For example, the LEAP algorithm~\citep{wolz2010leap} first learns a manifold structure on all (novel plus training) images. 
Next, a small number of novel images closest to the atlases are automatically segmented via a multi-atlas procedure.
These automatically segmented novel images are then added to the atlas list and the whole procedure is repeated. 
By using stepping stones, this strategy
boosts the performance in cases where some novel images are considerably different from the atlases.
\rev{There are other algorithms that also rely on self-training, i.e., using the automatically segmented images as new training data~\citep{chakravarty2013performing,liao2013sparse,shen2010supervised,wang2013groupwise}. These approaches have the disadvantage that segmentation mistakes reinforce themselves. Another way of exploiting unlabeled data is to use unlabeled scans to generate multiple deformations of a single labeled atlas to a novel scan, again using the unlabeled volumes as stepping stones~\citep{gass2013semi}. }

\subsection{Label Fusion}  

Label fusion, i.e., the step of combining propagated atlas labels, is one of the core components of MAS. 
The earliest and simplest fusion methods are best atlas selection~\citep{rohlfing2004evaluation} and majority voting~\citep{heckemann2006automatic, klein2005mindboggle, rohlfing2004evaluation}. 
In best atlas selection, a single atlas is utilized, which is usually chosen based on examining the match between the registered atlas and novel image intensities, for example, as captured by the registration cost function
(e.g., sum of squared differences, normalized cross-correlation, or mutual information). 
\rev{Relying on a single atlas disregards potentially useful information in all other atlases.}
Majority voting chooses the most frequent label at each location,
\rev{therefore using information from all atlases at all locations; however, it has the drawback that it ignores image intensity information}. 

An extension of majority voting is weighted voting, where each atlas is associated with a weight
(global or local) that reflects the similarity between the atlas and novel image.
The first method using global weights was proposed by \citet{artaechevarria2008efficient}, who used 
weights proportional to the normalized mutual information between the registered atlas image and novel image intensities. 
A related approach is to estimate the weights by posing it as a least squares problem, where the novel image intensities are assumed to be equal to the weighted combination of atlas intensities~\citep{cao2011putting}.
An alternative strategy involves defining the weights based on the similarity of the labels, which can be computed iteratively with respect to the current segmentation, either globally~\citep{langerak2010label} or within predefined ROIs~\citep{langerak2011local}; or estimated by examining the \rev{pairwise similarities} between the atlases~\citep{datteri2011estimation}.  

\rev{Global weights  cannot model the spatially-varying nature of registration accuracy. For this reason, the} 
use of global weights was later replaced by local and semi-local weighting schemes. 
The earliest examples of this strategy used weights inversely proportional to the absolute difference between local intensities of the novel image and deformed atlas~\citep{isgum2009multi,iglesias2009robust}, and standard local intensity-based registration metrics such as local cross-correlation~\citep{artaechevarria2009combination}. 
Alternative local weighting strategies were further explored, including the use of a precomputed local reliability measure~\citep{wan2008automated}, the Jacobian determinant of the deformation fields~\citep{ramus2010construction},  a Gaussian intensity difference function~\citep{depa2010robust, jia2012iterative}, the inverse of the squared standard score~\citep{tamez2012unsupervised}, a measure of the saliency of each atlas~\citep{ou2012attribute}, local mutual information~\citep{nie2013automated}, \rev{estimates of local registration accuracy~\citep{datteri2014applying}, and structural relationships between locally extracted wavelet features~\citep{kasiri2014cross}. }
Other studies have used weights defined as a function of ranks of local similarity, computed with correlations~\citep{yushkevich2010nearly} or Jacobian determinants~\citep{doshi2013multi}.  
Bridging global and local weighting, \citet{wolz2013automated} used weights that combined three different terms, reflecting global, organ-level and local (intensity-based) similarities. 
In a related effort,  a combination of region-wise and voxel-wise similarities \rev{(all based on sum of squares)} were used in~\citep{xie2014low}. 
\rev{Even though these strategies improve the segmentation accuracy obtained with global weights, the optimality of the chosen local weight metrics remains unclear.} 

In a series of papers, \citet{wang2011regression,wang2013multi} computed fusion weights that exploited the correlation structure between the atlases. \rev{The weights were optimized to minimize the expectation of segmentation error, which in turn led to relaxing the common independence assumption on the atlases}. Moreover, the registration-determined correspondence was refined via a local patch search.  
In a later paper, the same authors improved their algorithm  to make the  segmentations of the novel images consistent, such that the automatic segmentations are recruited as atlases, but with a lower weight than the manually labeled ones~\citep{wang2013groupwise}.

Other works have used more complicated schemes to define local weights, for example via offline learning.
One such method  assumed that the weights were a linear combination of the dissimilarities of the voxels at each location, and learned them with Tikhonov-regularized least squares (ridge regression)~\citep{khan2011optimal}.
Another related approach pre-registered all the atlases with each other to compute a reliability metric as the average agreement of the propagated labels; the reliabilities were then used as weights in the fusion~\citep{sdika2010combining}.
Along a similar direction, \citet{zhang2011confidence} used a forward-backward, patch-based search to compute a measure of correspondence specificity with respect to each atlas. 
Label fusion is then conducted in a sequential manner, starting at voxels that the algorithm is confident about segmenting and employing already segmented voxels within the neighborhood for guiding the segmentation of yet-to-be-labeled voxels. 
\rev{A related method that was recently proposed by \citet{koch2014graph}, uses a graph that connects similar regions across images to allow label information to iteratively flow from high confidence to low confidence voxels.}
In a different approach, \citet{wachinger2012spectral} used spectral clustering to identify homogeneous regions, and then performed semi-local label fusion within each region to finally compute a single label per region by pooling the votes within its boundaries. 

An alternative label fusion strategy involves the use of patches to compute weights at each voxel, which can be used with a conventional label fusion method~\citep{coupe2011patch, fonov2012multi}.
\rev{This technique has recently become more sophisticated. For example, \citet{xiao2014patch} and \citet{wang2014integration} used this approach to compute local label fusion weights using multi-channel MRI data.
\citet{wang2014geodesic} proposed to use the anatomical context to improve the quality of the patch matches.}
Instead of labeling the central voxel, one can segment the whole patch, and overlapping segmentations can then be fused (e.g. via majority voting)~\citep{rousseau2011supervised,sanroma2014novel}.
\rev{These methods have produced state-of-the-art segmentation accuracy, often at a high computational cost.}

Rather than directly using the similarity between patches, one can also compute the label fusion weights by seeking sparse linear combinations from a patch dictionary to reconstruct each patch of the novel image~\citep{liao2013sparse, zhang2012sparse,wang2014integration}.
\rev{Along a similar direction, \citet{sanroma2014novel} recently formulated label fusion as a matrix completion problem, which can be viewed as unifying the weight estimation framework with a learning-based approach.}
\Citet{cao2011segmenting} also used weights that best reconstructed the intensities of the novel images from the $k$ nearest atlases, computed on an image manifold. 
Instead of focusing on reconstruction error, a different method~\citep{wu2014generative} involves modifying the framework to reflect the consistency in the segmentations, such that  atlases that propagate similar labels to the segmentation have a similar contribution.

\rev{A different view of label fusion formulates segmentation as an optimization problem, where the agreement with the propagated atlases makes up a data fit term. In this framework, one can incorporate prior expectations such as spatial and temporal smoothness in longitudinal data, as in~\citep{li2014simultaneous}.}

Some of the label fusion techniques discussed above can be derived from probabilistic models of the data. 
Casting a segmentation method as a Bayesian inference problem in a probabilistic model has several advantages. 
First, it can easily deal with missing data, e.g., lack of labels in a given region of an atlas. 
Second, the estimated parameters of the model might have a direct interpretation that can provide us with some insight about the data and the fit of the model.
Third, the modeling assumptions have to be clearly stated and their effect on final accuracy can be empirically examined.
Fourth, the impact of the inference or estimation strategy can also be assessed by investigating alternative methods.
Finally, Bayesian methods are based on a principled and flexible framework, which can be adapted to the specifications of the problem at hand.

A generative probabilistic model of label fusion was first proposed by \citet{sabuncu2010generative,sabuncu2009supervised}. 
The model comprises of an \textit{unknown} discrete membership field 
that indexes the atlas that ``generated'' each voxel of the novel image and an additive Gaussian noise component.
This generative model framework unifies some of the most popular label fusion algorithms, generalizing local, semi-local and global weighted fusion methods, including majority voting and best atlas selection. 
The generative model has been extended to intermodality fusion~\citep{iglesias2013probabilistic}, replacing the Gaussian noise by a joint histogram; and to patch-based fusion~\citep{bai2013probabilistic}, by augmenting the membership field with a spatial shift and defining the intensity likelihood term as a function of patches.

\rev{There is a family of generative models for label fusion} that can be viewed as a modification of  Sabuncu's model~\citep{sabuncu2009supervised, sabuncu2010generative}, where the latent membership field is only used to define a prior on the labels and the novel image intensities are generated directly from the underlying segmentation, e.g., via a parametric Gaussian.
This model does not utilize the relationship between the image intensities and labels observed in the atlases and thus can be used to segment images of a modality different from the atlases~\citep{iglesias2012isbi}, or multi-channel images~\citep{iglesias2012generative}. \rev{This strategy will be suboptimal for scenarios where the intensity profiles of the atlases and the novel scans are matched}.
Iglesias, Tang and colleagues later proposed to integrate registration into this generative model~\citep{iglesias2013unified, tang2013bayesian},
\rev{which offers a small but significant improvement in segmentation accuracy at an increased computational cost}. 
Finally, many methods that use label fusion to construct a prior in a probabilistic segmentation algorithm~\citep{lotjonen2009atlas, van2008hippocampus,van2012automated, wang2014automated,wachinger2014atlas,wolz2009segmentation,wolz2010measuring,platero2014multiatlas} can be viewed to be (approximate and/or modified) instantiations of the probabilistic generative label fusion framework.

Another family of probabilistic fusion methods builds on the STAPLE algorithm~\citep{warfield2004simultaneous}. 
STAPLE was originally developed to model manual segmentations as noisy observations of the hidden (ground truth) segmentation and the noise was modeled with a stationary confusion matrix $\{\theta_n\}$. 
The original STAPLE algorithm only supported binary segmentations~\citep{warfield2004simultaneous}, but was soon after extended to the multi-class setting~\citep{rohlfing2003expectation,rohlfing2003expectation2,rohlfing2003extraction}. 
Many extensions of STAPLE correspond (or can be shown to correspond) to modifications of the original probabilistic model, for example placing a Beta prior on the parameters of the confusion matrix~\citep{commowick2010incorporating}, replacing the hard atlas segmentations with probabilistic maps~\citep{weisenfeld2011softstaple}, dealing with missing atlas label data~\citep{landman2012robust}, altering the confusion matrix to account for self-assessed uncertainty~\citep{asman2011robust,bryan2014self}, employing a hierarchical noise model~\citep{asman2014hierarchical}, {\rev{introducing and estimating unknown reliability weight maps~\citep{akhondi2014logarithmic}, and learning and exploiting the relationship between performance parameters and intensity similarities~\citep{gorthi2014optimal}}.

Rather than making explicit changes to the original framework and solving the corresponding model, some extensions of STAPLE are based on \textit{ad-hoc} modifications.
For instance, some researchers have introduced spatially varying performance parameters to the model by estimating local confusion matrices from windows around each voxel~\citep{asman2012formulating,commowick2012estimating}. 
One can view these methods as approximate solvers to a version of the STAPLE model, in which the noise parameters vary smoothly over space. 
\rev{In a different approach, \citet{langerak2010label}, \citet{cardoso2013steps}, \citet{nouranian2014multi}, and \citet{weisenfeld2011learning} proposed using only a subset of atlases in label fusion. Langerak \etal's SIMPLE algorithm integrates the atlas selection step into STAPLE and solves for that iteratively. The SIMPLE method was recently integrated with context learning to exploit exogenous information, e.g, about tissue likelihood~\citep{xu2014simple}. Cardoso \etal's method obtains the subset by ranking the atlases in terms of local similarity to the novel image; Nouranian \etal's algorithm iteratively computes the segmentation with STAPLE and removes the atlases that do not agree with the current estimate of the labels; and Weisenfeld \etal use a probabilistic formulation to disregard atlases that do not agree with the current segmentation estimate. Again, these algorithms} can be seen as an approximate solution to a model, in which the atlases to be explained are indicated by a latent field.
Finally, \citet{asman2013non} incorporated information from intensity image patches in STAPLE. 
From a probabilistic modeling perspective, this approach would require modifying STAPLE's model to connect the novel image intensities to the training images.

\subsection{Post-processing}  

The label fusion result does not necessarily represent the final segmentation; sometimes it is fed to another algorithm to estimate the output labels. 
The extent to which this post-processing changes the segmentation varies across methods. 

Some methods use the output of label fusion to simply initialize a subsequent algorithm, for instance, to determine the bounding box where a segmentation method is applied~\citep{van2007automatic}, to start the evolution of an active contour~\citep{fritscher2014automatic, hollensen2010segmenting}, {\rev{or to fit a smooth contour to the object boundary~\citep{nouranian2014multi}}.
Other MAS algorithms rely on applying heavy post-processing to the label fusion output, for example by employing an error detection and correction classifier (\citealt{yushkevich2010nearly}, who use AdaBoost), deriving features to drive a subsequent voxel-wise segmentation method, \rev{based for example on level sets~\citep{gholipour2012multi,schreibmann2014multiatlas}, random forests~\citep{han2013learning}, support vector machines~\citep{hao2014local}, patch-based techniques~\citep{wang2014segmentation}, or a graph-cut-based method~\citep{candemir2014lung,lee2014fully}}.
\rev{Along a similar direction, one can apply a refinement to the MAS results, for example, by comparing the observed intensities in the novel image to tissue-based expected intensity profiles~\citep{ledig2014robust}}. 
Alternatively, label fusion results have been used to compute priors in probabilistic segmentation algorithms~\citep{fortunati2013tissue, shi2010construction,van2008hippocampus, van2012automated, wang2014automated, wolz2009segmentation, wolz2010measuring, xu2014shape, platero2014new, makropoulos2014automatic,yan2014accurate}. \rev{These methods can be robust to changes in image appearance, for example, in applications where there is significant variation in imaging parameters or the novel subject's anatomy is not represented in the atlases. However, they will be less accurate than standard MAS methods when the intensity profiles and appearance distribution are well matched between the novel image and atlases. In a related effort, \citet{liu2014detection} used MAS to define a prior for the detection of lymph nodes in thoracic CT scans.}
A different strategy is to examine summary measurements (e.g., volume of an ROI) computed from the MAS to statistically determine whether the segmentation result is an outlier and thus might have failed -- in which case one can resort to manual 
delineation~\citep{van2009automatic2}.

There are also methods that operate on the posterior probability map obtained from label fusion, rather than applying a hard threshold to obtain a segmentation.
For example, applying a deconvolution to the probability map has been shown to reduce the spatial bias in the segmentation of convex structures~\citep{wang2012spatial}.
In the context of neointima (scar tissue) segmentation in coronary optical coherence tomography, \citet{tung2013multi} augment the posterior probability  with an anatomically-informed probability, \rev{defined upon the distance to the vessel wall}. \rev{While this prior knowledge enhances the performance of the method, it is highly domain specific and not applicable to other problems.}
In a parallel approach,  \citet{asman2013out} propose to analyze the posterior probabilities to detect \rev{outliers} that are not well represented in the atlas set. \rev{This is shown to be beneficial in the presence of anomalous regions (e.g., tumors).}

\section{Survey of Applications}
\label{sec:app}

Since its original application to confocal microscopy of bee brains~\citep{rohlfing2004evaluation,rohlfing2005multi}, MAS has been successfully used in a large variety of biomedical segmentation problems. 
The most prevalent field of application has been brain MRI analysis, for two different reasons; first, segmentation's crucial role in a wide range of widely studied neuroimaging problems; and second, the success of image registration techniques in this field. 

Most of the MAS work applied to brain MRI data has focused on the segmentation of cortical and subcortical regions in structural images, typically acquired with T1-weighted MRI sequences. Many methods have been developed to parcellate the whole brain, segmenting it into a large number of regions~\citep{aljabar2008automated,babalola2009evaluation,fonov2012multi, han2009gpu, heckemann2010improving, heckemann2011automatic, keihaninejad2010automatic, kotrotsou2014ex,svarer2005mr,wang2012groupwise,ledig2014robust}, while other studies have focused on small sets or individual ROIs, such as the caudate nucleus~\citep{van2007multi}; the cerebellum~\citep{park2014derivation,van2012automated,weier2014rapid}; the amygdala~\citep{hanson2012robust, klein2014amygdalar}; the corpus callosum~\citep{ardekani2014corpus,gao2014multi,meyer2014multi}; \rev{the striatum~\citep{janes2014striatal}};  \rev{the subthalamic nucleus, red nucleus and substantia nigra~\citep{xiao2014investigation,xiao2014patch}}; the ventricles~\citep{chou2008automated,raamana2014three}; and, most notably, the hippocampus, which has attracted much attention due to its association with dementia and Alzheimer's disease~\citep{akhondi2010multiple, bishop2010evaluation, clerx2013measurements, hammers2007automatic, iglesias2010synthetic, kim2012automatic, leung2010automated,  pipitone2014multi, pluta2012vivo, raamana2014three, van2008hippocampus, van2012automated, winston2013automated, wolz2010measuring, yushkevich2010nearly,platero2014new,ta2014optimized}.

In the context of segmentation of structural human brain MRI, multi-atlas techniques have also been applied to preprocessing tasks such as skull stripping~\citep{leung2011brain, weisenfeld2011softstaple} and tissue classification~\citep{bouix2007evaluating,crum2009spectral}, the segmentation of tumors~\citep{zikic2014classifier,wang2013multi2,warfield2004simultaneous}, \rev{eyes and optic nerves~\citep{datteri2014applying,harrigan2014robust}}. 
\rev{MAS has also been employed for the segmentation of cortical and subcortical structures in MRI data from fetuses, neonates, and infants too~\citep{gholipour2012multi,gousias2008automatic,gousias2010atlas,gousias2013magnetic,shi2010construction,wang2014segmentation,li2014simultaneous,koch2014graph, makropoulos2014automatic,wang2014integration}, in which the contrast inversion due to ongoing myelination complicates the segmentation.}
Another area of application of MAS has been the segmentation of brain MRI in animal studies, e.g., mice~\citep{ma2012multi,ma2014automatic, nie2013automated,lee2014multi,khan2014molecular}, \rev{rats~\citep{lancelot2014multi}} and non-human primates~\citep{ballanger2013multi}. 
Finally, there are also studies that have applied MAS to the analysis of diffusion brain MRI data of humans~\citep{jin2012automatic,tang2014multi,traynor2010reproducibility}, which requires specific strategies for the registration, atlas selection and label fusion steps, due to the nature of the data, which are typically described by directional functions defined on the sphere at each voxel.

Outside brain imaging, the prevalence of prostate cancer in men has sparked interest in applications within prostate imaging, using modalities such as MRI~\citep{langerak2010label,litjens2014computer,rivest2014fast}, CT~\citep{acosta2011evaluation,sjoberg2013clinical,acosta2014multi} and ultrasound~\citep{nouranian2014multi}. 
Likewise, interest in radiotherapy treatment planning has been the main driver of applications in head, neck, \rev{and thoracic} CT segmentation~\citep{han2008atlas,wang2014automated}, which have mainly focused on segmenting tumors~\citep{ramus2010assessing}, organs at risk (e.g, the parotid glands, \citealt{fritscher2014automatic, gorthi2010multi, han2010automatic, hollensen2010segmenting, yang2010automatic} \rev{or mediastinal lymph nodes, \citealt{liu2014detection}}) and lymph node metastases~\citep{sjoberg2013clinical, teng2010head}. 
MAS has also been used in abdominal imaging, despite the relatively poor performance of image registration in this domain (e.g., compared with brain MRI) due to the shifting of organs within the abdominal cavity. 
Nonetheless, MAS has been successful in liver~\citep{van2007automatic,platero2014multiatlas}, spleen~\citep{li2013regression,xu2014shape} and multi-organ segmentation~\citep{wolz2013automated,schreibmann2014multiatlas} in CT scans.

Finally, there are many other applications that have benefited from MAS within human medical imaging, including: segmentation of pelvic bones in MRI~\citep{weisenfeld2011softstaple,akhondi2014logarithmic}; lungs in CT scans~\citep{van2009automatic2} and \rev{chest X-rays~\citep{candemir2014lung}}; heart and its ventricles in CT~\citep{van2010adaptive,dey2010automated}, MRI~\citep{zhuang2010whole, zuluaga2014multi}, MR angiography~\citep{wachinger2012spectral}, \rev{ultrasound~\citep{wang2014multiLBLF}}, and CT 
angiography~\citep{kiricsli2010evaluation, yang2014automaticISBI}; breast tissues and lesions in X-ray mammography~\citep{iglesias2009robust} and MRI~\citep{gubern2012segmentation,lee2013breast}; cartilage and bone in knee MRI~\citep{tamez2012unsupervised,lee2014fully,shan2014automatic}; the vertebrae in spinal MRI~\citep{asman2014groupwise}; scar tissue in intravascular coronary optical coherence tomography (OCT)~\citep{tung2013multi}; the mitral valve in transesophageal echocardiography~\citep{wang2013multi3,pouch2014fully}; \rev{skeletal muscle in whole-body MRI~\citep{karlsson2014automatic}; kidneys in CT images~\citep{yang2014automaticEMBC}; and bone in dental cone-beam CT images~\citep{wang2014automated}.}

\section{Discussion and Future Directions}
\label{sec:disc}


By taking full advantage of the entire training data, rather than a model-based summary, MAS delivers highly accurate segmentation algorithms.
This approach has come a long way since the early days of ``majority voting'', which basically consisted of three independent steps: registration, label propagation, and fusion.
Today, most MAS algorithms have many more steps, some of which form feedback loops. 
Furthermore, each one of these steps is becoming increasingly more sophisticated, employing ideas from optimization, computer vision, machine learning, probabilistic modeling, and other fields.

The biggest shortcoming of MAS is its ravenous appetite for computational resources.
Analyzing, manipulating, and processing all atlases typically demands a substantial amount of memory and time.
\rev{We believe this is one of the main reasons why MAS has not been widely adopted in clinical applications yet, even though, research suggests that it can produce state of the art segmentation tools in many domains.}
However we expect that several recent developments alleviate 
\rev{the computational challenges of MAS.}
Firstly, the continued exponential growth in computer hardware technologies is to our advantage.
We note, however, this technological benefit is to some extent countered by the rapidly increasing resolution of biomedical images, which multiplies the computational burden.
Secondly, we observe that many of the subcomponents of MAS are parallelizable and thus can take advantage of multi-core architectures and GPUs. 
At the coarsest level, the registrations that need to be computed with each atlas can be solved in parallel.
Furthermore, each registration can be implemented such that the bulk of the voxel- or region-level computations can be distributed over multiple processors.
This approach has already been used for the GPU-acceleration of the registration step~\citep{cardoso2013steps, duc2013using,han2009gpu,modat2010fast}.
A similar strategy can be adopted in the label fusion step, particularly by algorithms that conduct numerical optimization in label fusion and not just simple counting.
Finally, some of the online computational burden can be shifted offline, via learning structure on the training images, which can then be utilized to optimize the processing of the novel image, as proposed in~\citep{jia2012iterative}.

The manually delineated training data form the main foundation of atlas-based segmentation.
Empirical evidence suggests that the number and quality of training cases can critically impact segmentation accuracy.
\rev{For this reason, the careful definition and standardization of annotation guidelines is paramount to obtaining accurate automatic segmentations, especially when the atlases are manually delineated by multiple experts.}
Yet, obtaining high quality segmentations annotated by experts is both time consuming and expensive.
Most past research has dealt with scenarios where the development of the segmentation algorithm is independent of the manual segmentation process.
We believe a better strategy is to integrate the two pipelines.
For example, as recently demonstrated~\citep{awate2014multiatlas}, given a segmentation method, one can estimate the number of cases that need to be manually delineated to achieve a desired level of accuracy.

Furthermore, one can imagine an algorithm that indicates the cases, which, if manually segmented, assist the segmentation algorithm the most.
Active learning can provide the framework to derive such an algorithm.
\rev{An alternative approach is to use automatic segmentations as atlases, after applying a quality control step.
Yet a different strategy is to harness the potential of non-expert segmenters~\citep{bogovic2013approaching,bryan2014self}, for example, via a crowd-sourcing framework~\revv{\citep{landman2012foibles,maier2014crowdsourcing}.}}
Although many biomedical segmentation problems rely on anatomical expertise, it is not clear whether this expertise has to be deployed in the delineation of every single atlas.
One can imagine certain scenarios, where the expert(s) provides a handful of example annotations, which can be used to train or guide non-experts.
Finally, we believe that the idea to combine heterogeneous sets of atlases, delineated with different protocols, is a promising future direction. 
This strategy can both yield better accuracy by enriching the training data and offer the ability to identify ROIs that were technically not part of any single manual delineation protocol but can be defined by intersections\rev{~\citep{iglesias2015algorithm}}. 
\rev{Moreover, such an approach could also potentially minimize the impact of the variability of the manual delineations (within or across experts) on the final segmentation, automatically learning the biases of the annotations.}
 
\rev{Crowdsourcing offers another attractive solution to the atlas generation problem of MAS. Instead of  high quality manual delineations from trained experts, one might consider using lower quality data from non-experts\revv{~\citep{maier2014crowdsourcing}}. Alternatively, the non-expert crowd can be used to correct or filter the segmentations. We expect that outsourcing certain aspects of MAS, particularly those related to the offline stages of the pipeline, to non-expert and/or expert masses in an online community will be investigated in the near future.} 
 
While speeding up the registration step might be considered top priority for some applications, many biomedical problems seek very high accuracy, even at high computational cost.
For such applications, one strategy is to improve registration accuracy and the quality of propagated labels. 
The probabilistic modeling perspective offers a complementary approach.
From this viewpoint, registration is a nuisance parameter and thus should be marginalized out, e.g., via variational techniques~\citep{simpson2011probabilistic} or a sampling procedure such as Markov Chain Monte Carlo~\citep{iglesias2013improved}. In other words, 
 one should integrate over all possible registration results, rather than attempting to find the most likely one and using that for the fusion step.
Note that this approach would be different from the latest algorithms that combine the registration and label fusion steps, as done in~\citep{hao2012iterative,iglesias2013unified,tang2013bayesian}. Currently, the marginalization strategy might seem computationally prohibitive for MAS.
However, the recent successful applications of this idea in other biomedical image analysis scenarios suggest that in the near future we can expect to see label fusion algorithms that integrate out the unknown registrations.

\rev{Another direction of future work in MAS is to develop algorithms that are robust against changes in image intensity profiles, e.g., MRI contrast, due to variation in acquisition protocols, hardware, and other imaging parameters. Such robust methods will be invaluable for large-scale multi-site studies and clinical applications, where the standardization of the imaging protocol might be unrealistic. Although some existing label fusion algorithms (e.g., \citealt{iglesias2012isbi,iglesias2013probabilistic}) have been developed to handle different modalities, they are application-specific and make strong assumptions about the data (e.g., locally Gaussian intensity distributions).}

We believe that the fields of machine learning and computer vision have also a lot to contribute to MAS.
Recent years have witnessed dramatic technical advances in both of these fields, such as unsupervised feature learning in vision~\citep{erhan2010does} and efficient learning methods on deep architectures~\citep{hinton2006reducing}, which have facilitated tremendous gains in performance.
Recent developments suggest that researchers are currently working on translating such ideas to biomedical image analysis problems, including MAS.

The probabilistic perspective, with its flexibility and principled inference machinery, offers another promising direction for future research. 
In particular, this approach enables the derivation of methods that can handle missing labels, heterogeneous labels, variable imaging modalities, estimate and utilize model uncertainty, and integrate domain knowledge, for example about the anatomy or imaging physics.
Furthermore, probabilistic algorithms offer the capability to quantify the uncertainty in the final segmentation estimate, which can further be utilized for obtaining more accurate measurements, for example of the volume of structures~\citep{iglesias2013improved}.

Rather than segmenting each novel image independently, empirical evidence suggests that solving the segmentations of multiple novel images simultaneously might yield improved results~\citep{wang2012groupwise}.
This can be a particularly promising approach for segmenting serial scans.
Longitudinal image analysis is an area of growing importance and the detection of subtle longitudinal changes can call for highly accurate segmentation~\citep{reuter2012within}.
Encouraged by some recent applications (\citealt{wolz2010measuring} \rev{and \citealt{li2014simultaneous}}), we believe MAS will be a critical tool for longitudinal biomedical image analysis.

So far, most of the applications of MAS have been in the domain of human brain MRI, in which modern registration algorithms achieve good alignment and even the simplest fusion algorithms (e.g., majority voting) yield good performance. Registration is however less effective in other modalities and body parts, such as in abdominal imaging, in which the sliding between organ walls (e.g., due to respiratory motion) is problematic for current algorithms. We believe, though, that the development of registration methods that can cope with these difficulties, along with the improvements in label fusion techniques (which will make them more robust against misregistration), will make the use of the multi-atlas approach ubiquitous in a growing number of novel biomedical image segmentation problems.

\rev{Finally, it is important to note that there is no \emph{universally} optimal segmentation algorithm. 
Each application brings with it a unique set of constraints and objectives, making certain types of methods more suitable than others. 
Yet, we believe that the large class of MAS methods, with its rich set of instantiations that enable compromising between different application tradeoffs and considering various objectives while exploiting different sources of information, offer a framework that promises to yield effective and useful solutions for a wide range of biomedical applications. 
That said, we can identify general trends that have emerged. 
There seems to be a global tradeoff between computational efficiency and segmentation accuracy.
Incorporating domain knowledge and adopting realistic models that are based on the physical and biological context, can yield significant improvements.
Complex, more advanced methods can pay off and should be something we continue to work on. 
However, this endeavor critically depends on a proper evaluation of the methods,  as demonstrated in some recent efforts~\citep{rueda2014evaluation,menze2014multimodal, panda2014evaluation,goksel2014segmentation}.
Going forward, a grand challenge of biomedical image segmentation will be to establish standardized datasets and performance evaluation metrics to be used to objectively compare various segmentation algorithms, including MAS-based techniques. } 

\section*{Acknowledgment}
Juan Eugenio Iglesias is supported by the Gipuzkoako Foru Aldundia (Fellows Gipuzkoa Program). 
Mert R. Sabuncu is supported by NIH NIBIB 1K25EB013649-01 and a BrightFocus Alzheimer's disease pilot research grant (AHAF-A2012333).

\bibliographystyle{model2-names}
\bibliography{bibliography_only_MAS_V3}
\end{document}